\documentclass{bmvc2k}

\def\thefootnote{*}\footnotetext{Equal contribution}

\usepackage{multicol}
\usepackage{multirow}
\usepackage{amsfonts}
\usepackage{booktabs}
\usepackage{xfrac}    % for \xfrac macro
\usepackage{nicefrac} % for \nicefrac macro
\usepackage{amsmath}  % for \dfrac macro
\usepackage{url}

\usepackage{paralist}
\usepackage[inline]{enumitem}

\makeatletter
\def\thickhline{%
  \noalign{\ifnum0=`}\fi\hrule \@height \thickarrayrulewidth \futurelet
   \reserved@a\@xthickhline}
\def\@xthickhline{\ifx\reserved@a\thickhline
               \vskip\doublerulesep
               \vskip-\thickarrayrulewidth
             \fi
      \ifnum0=`{\fi}}
\makeatother

\newlength{\thickarrayrulewidth}
\title{Few-Shot Domain Adaptation for Low Light RAW Image Enhancement}

\addauthor{K Ram Prabhakar}{ramprabhakar@iisc.ac.in}{1}
\addauthor{Vishal Vinod\thefootnote{}}{vishal114186@gmail.com}{1}
\addauthor{Nihar Ranjan Sahoo\thefootnote{}}{niharsahooigit@gmail.com}{1}
\addauthor{R. Venkatesh Babu}{venky@iisc.ac.in}{1}

\addinstitution{
 Video Analytics Lab,\\
 Department of Computational and Data Sciences,\\
 Indian Institute of Science,\\
 Bangalore, India
} 
\runninghead{Ram, Vishal, Nihar, Babu}{FSDA for Low Light RAW Enhancement}

\begin{document}

\maketitle

\begin{abstract}
Enhancing practical low light raw images is a difficult task due to severe noise and color distortions from short exposure time and limited illumination. Despite the success of existing Convolutional Neural Network (CNN) based methods, their performance is not adaptable to different camera domains. In addition, such methods also require large datasets with short-exposure and corresponding long-exposure ground truth raw images for each camera domain, which is tedious to compile. To address this issue, we present a novel few-shot domain adaptation method to utilize the existing source camera labeled data with few labeled samples from the target camera to improve the target domain's enhancement quality in extreme low-light imaging. Our experiments show that only ten or fewer labeled samples from the target camera domain are sufficient to achieve similar or better enhancement performance than training a model with a large labeled target camera dataset. To support research in this direction, we also present a new low-light raw image dataset captured with a Nikon camera, comprising short-exposure and their corresponding long-exposure ground truth images. The code is available at \url{https://val.cds.iisc.ac.in/HDR/BMVC21/index.html}.
\end{abstract}

%-------------------------------------------------------------------------
\section{Introduction}
\label{sec:intro}
    Capturing high-quality photos in low illumination is a fundamental yet challenging task. Increasing the ISO improves visibility; however, it also increases sensor noise. Longer exposure times improve the image but require a tripod to avoid camera motion and motion blur. Methods like enabling flash or image editing also present their own challenges. Low exposure image enhancement helps generate low-light scenes as if they were captured with a longer exposure time. It enables fast low illumination photography without a tripod. As shown in Fig. \ref{fig:canon_eg1}(a), the images captured in such settings possess a high degree of noise and color distortion. Existing single image denoising methods \cite{dabov2007image, plotz2017benchmarking} perform poorly and fail to correct color distortions in low light. 
\newpage 
An alternative is to merge a burst of short exposure images to reduce noise \cite{hasinoff2016burst,liu2014fast,mildenhall2018burst}. However, the burst images must be aligned for the camera and object motion, and aligning them in low-light conditions is a challenge. LSID, \cite{chen2018learning} a recent deep learning based method enhances raw low-light images by performing denoising, color improvement, and demosaicing, all with a single lightweight model. Despite the success of recent methods, there are two persisting challenges:

\begin{figure*}[t]
    \centering
    \label{fig:canon_eg1}
    \includegraphics[width=\linewidth, clip, trim=0cm 0.75cm 0cm 0.661cm]{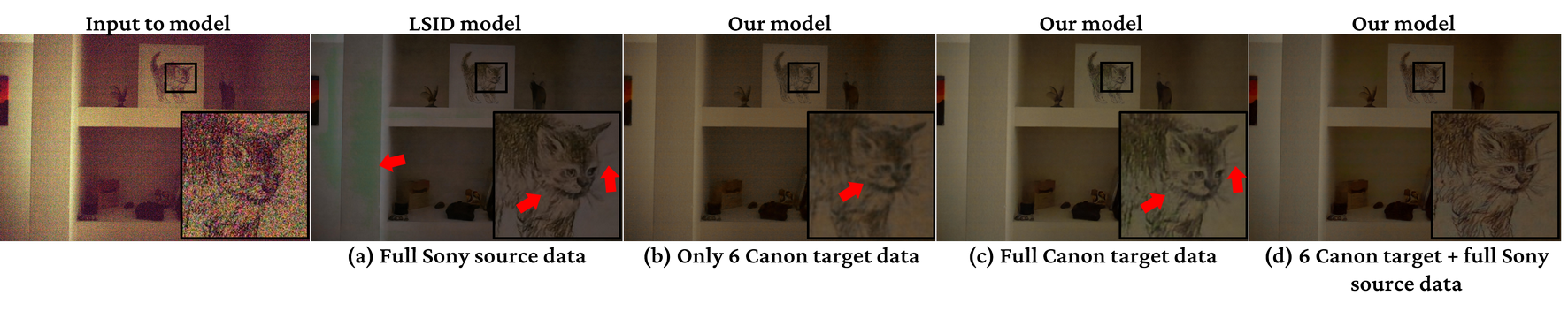}
    \caption{Qualitative comparison between different methods with an input to the model from the Canon dataset (left-most). The models are: LSID trained on (a) full 161 Sony source images, (b) only 6 Canon images, and (c) full Canon dataset. (d) 6 Canon images and 161 Sony images with proposed few-shot domain adaptation method (discussed in section \ref{sec:method}).}
    \vspace{-0.3cm}
\end{figure*}

%Deep learning-based techniques generate high-quality results as compared to traditional non-learning methods and their success can be attributed to the large labeled train data in the raw domain. Raw data is uncompressed and minimally manipulated information directly from the digital camera sensors. Thus, raw data captured with a camera from one manufacturer differs from that of another camera manufacturer. Many cameras have their own proprietary raw image format to save sensor data.

% thus have different 'shot' noise (photon arrival statistics) and 'read' noise (readout imprecision) \cite{hasinoff2014photon} at high ISO amplification.

\textbf{Domain Shift}: CNNs are heavily data-sensitive; Since raw images captured with cameras from different manufacturers exhibit variations in color-space and noise characteristics, a model trained with one camera's raw data performs sub-optimally on another camera's raw data (see Table \ref{tab:baseline}). Hence, there exists a domain shift across different camera raw domains. In this work, we consider each camera as a separate domain. As seen from the Fig. \ref{fig:canon_eg1}, cross-camera domain performance is poor as there are color distortions (green patches on the wall) and loss of finer details (missing cat's whiskers) due to the shift across camera domains. %As we show in Table \ref{tab:baseline}, an LSID model trained on Nikon camera data performs sub-optimally on Sony and Canon camera datasets. Similarly, a model trained on Canon camera data performs well only for Canon raw images Fig. \ref{fig:canon_eg1} (b) shows an example of a model trained with Sony camera raw data and tested on Canon camera raw data. 

\textbf{Tedious to collect labeled data}: Collecting a large dataset of short-exposure and long-exposure raw image pairs for each camera is a difficult task. There must be no object motion, and to avoid camera misalignment, a tripod is necessary to capture long-exposure images (typically 10 seconds or more). Further, a smartphone or an IR remote is required to trigger the camera to avoid camera shake arising from physically pressing the camera click button. Thus, capturing a large-scale labeled dataset for different cameras is an arduous task.

To address the above mentioned challenges, we propose a paradigm shift for the low-light raw image enhancement task using few-shot learning and domain adaptation. We use a large collection of existing source camera labeled data to improve the performance and generate the output in the target domain by transferring the task onto a new target camera dataset with only a few labeled samples. In summary, our contributions are as follows:
% We address these challenges by designing a novel, simple, and lightweight approach for the low-light raw image enhancement task by adapting from an existing camera domain with large labeled data (source domain) to a new camera domain with few labeled data (target domain). In this work, we propose a paradigm shift for the low-light raw image enhancement task in few-shot learning and domain adaptation. We use a large collection of existing source camera labeled data to improve the performance and transfer the task onto a new target camera dataset with only a few labeled samples and generate the output in the target domain. In summary, our contributions are as follows:
% \begin{itemize}
\begin{enumerate}[label=(\roman*),wide,itemindent=1em,noitemsep,topsep=0pt,itemsep=0pt]
\item To the best of our knowledge, we propose the first few-shot domain adaptation method for low-light raw image enhancement. 
\item We show that, with less than ten labeled samples from the target domain, our approach can outperform a model trained with a complete target domain dataset. 
\item We present experiments and ablations to illustrate the effectiveness of our method.
\item We present a new Nikon camera dataset with short-exposure and long-exposure ground truth raw image pairs for the benefit of the research community.
% \end{itemize}
\end{enumerate}

% % Please add the following required packages to your document preamble:
% \begin{table}[t]
% \centering
% \setlength\tabcolsep{2pt}
% \caption{Comparison between LSID models trained with one source camera dataset and tested on other target camera datasets.}%\cite{chen2018learning}
% \label{tab:baseline}
% \scalebox{0.85}{
% \begin{tabular}{lcccccc}
% \hline
% Testing ($\rightarrow$) & \multicolumn{2}{c}{Sony} & \multicolumn{2}{c}{Nikon} & \multicolumn{2}{c}{Canon} \\ \hline
% Training ($\downarrow$) & PSNR & SSIM & PSNR & SSIM & PSNR & SSIM \\ \hline
% Sony  & \textbf{28.50  } & \textbf{0.774}   & 25.90  & 0.693   & 27.41   & 0.845   \\
% Nikon  & 19.95   & 0.481   & \textbf{30.74 } & \textbf{0.913}  & 24.34  &   0.767  \\
% Canon  & 18.51   & 0.542   & 23.27   & 0.847   & \textbf{32.32  } & \textbf{0.899}   \\ \hline
% \end{tabular}
% }
% \end{table}

%--------------------------------------------------------------------------------------------------------------------

\vspace{-0.5cm}\section{Related Work}
\label{sec:related}
    \begin{figure*}[t]
\centering
\scalebox{0.9}{
\subfigure{\includegraphics[width=\linewidth]{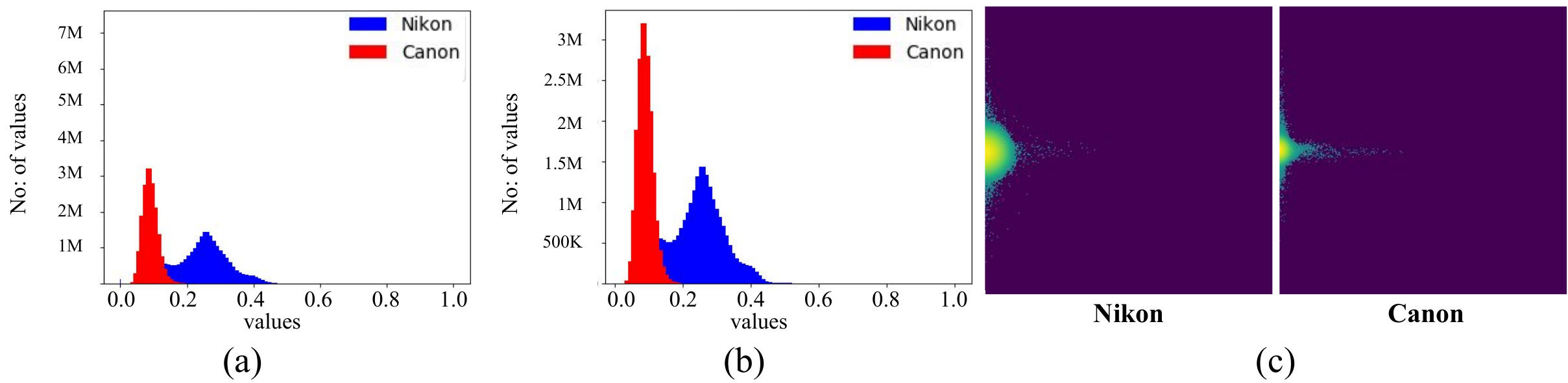}}
}
\caption{First-order and second order statistics for an image captured with Nikon and Canon cameras. (a) Histogram of intensities, (b) Histogram of spatial derivatives of intensity, (c) Joint histogram of responses from convolution filters.}
\label{fig:canon136}
\vspace{-0.4cm}
\end{figure*}

\begin{table}
\footnotesize
\centering
\setlength\tabcolsep{3pt}
\begin{minipage}[b]{0.52\textwidth}
\centering
\caption{Comparing LSID models trained with a complete source camera dataset and tested on respective target camera datasets.}%\cite{chen2018learning}
\label{tab:baseline}
% \tablewidth=\textwidth
\scalebox{0.95}{
\begin{tabular}{lcccccc}
\hline
Testing ($\rightarrow$) & \multicolumn{2}{c}{Sony} & \multicolumn{2}{c}{Nikon} & \multicolumn{2}{c}{Canon} \\ \hline
Training ($\downarrow$) & PSNR & SSIM & PSNR & SSIM & PSNR & SSIM \\ \hline
Sony \cite{chen2018learning}  & \textbf{28.50  } & \textbf{0.774}   & 25.90  & 0.693   & 27.41   & 0.845   \\
Nikon  & 19.95   & 0.481   & \textbf{30.74 } & \textbf{0.803}  & 24.34  &   0.767  \\
Canon \cite{CanonLSID}  & 18.51   & 0.542   & 23.27   & 0.847   & \textbf{32.32  } & \textbf{0.899}   \\ \hline
\end{tabular}
}
% \caption{This is a ver very very long caption which doesn't overwrites the text on the right side of the paper.}
% \label{tab:accuracy} 
\end{minipage}%
\hfill
\begin{minipage}[b]{0.42\textwidth}
\centering
\caption{Details of the datasets used in our work.}
\label{tab:datasets}
\scalebox{0.95}{
\begin{tabular}{lccc}
    \hline
    \multirow{2}{*}{Datasets} & Exposure & Training & Testing \\
    & Ratios & Images & Images \\
    \hline
    Sony \cite{chen2018learning} & 90,15,300 & 161 & 36\\
    Nikon & 100,300 & 53 & 24\\
    Canon \cite{CanonLSID} & 50,150,300 & 44 & 21 \\ \hline
    \end{tabular}
}
%  \caption{Speed up for the parallel solution of the trivial problem, 16
% Threads on Dual Xeon E-2690.} 
%  \label{tab:ompdiff}
\end{minipage}
\vspace{-0.5cm}
\end{table}

\textbf{Low-light Image Enhancement}: Existing low-light image enhancement methods require paired low/well-lit image scenes in RGB space and assume the images to be captured with minimal noise. While such methods capture global information suitably, their performance in extreme low-light conditions is sub-par. The histogram equalization method is useful for increasing the dynamic range in a global context and is sub-optimal for extreme low-light enhancement \cite{chen2018image}. Retinex \cite{park2017low} methods assume that the images have sufficient information to map the reflectance and enhance the low-light image. Similarly, \cite{Wang_2019_CVPR} formulates the illumination estimation for enhancing underexposed images akin to expert retouched ground truth without image-to-image regression. Methods such as EnlightenGAN \cite{jiang2021enlightengan}, a generative model with an attention U-Net \cite{ronneberger2015u}, \cite{xu2020learning} for decomposition and enhancement and MIRNet \cite{Zamir2020MIRNet} for attention aggregation have performed well in sRGB low-light enhancement. %While there are interesting developments in the HDR domain \cite{wu2018deep, cai2018learning}, the works are different from enhancing static low-light images. While such works have shown improved enhancement, their performance and relevance to extreme low-light images with significant noise and color distortion are yet to be explored in the raw image domain. 

\textbf{Single Image Denoising}: Non-deep methods for single image denoising either use engineered features \cite{simoncelli1996noise, rudin1992nonlinear}, or assume the noise model to be uniform and additive. Such assumptions in parametric methods are unsuitable for real-world low-light enhancement. Non-parametric methods depend on sparse image priors such as smoothness and self-similarity \cite{gu2014weighted, mairal2009non, loza2013automatic, dabov2007image} and are more expressive than parametric methods. BM3D \cite{dabov2007image, plotz2017benchmarking}, a non-blind denoising method requiring noise and color model information, has outperformed deep methods in accuracy and noise robustness; however, it is prone to over smoothing in low light conditions. Several deep methods leveraging CNN based advancements have been proposed for denoising \cite{zhang2017beyond, jain2008natural, xie2012image, zhang2018ffdnet, ulyanov2018deep}. Autoencoders \cite{xie2012image, lore2017llnet} and MLPs \cite{burger2012image} have shown sub-par performance on real-world raw sensor noise. Unprocess \cite{Brooks_2019_CVPR} learns the denoising pipeline and relevant photometric parameters by `unprocessing' the image and is different from our goal to learn the camera's color and noise model effectively.
% BM3D performs non-blind denoising requiring noise and color model information and is also prone to over-smoothing in low-light. 
% Most deep learning based methods have been evaluated on synthetic random noise rather than on real-world data. Joint denoising and demosaicing \cite{gharbi2016deep} are designed for general noise models and are not well suited for noise in the raw domain.

% \textbf{Deep Learning Methods.} Several deep learning methods leveraging Convolutional Neural Network (CNN) based advancements have been proposed for denoising and image enhancement \cite{zhang2017beyond, jain2008natural, xie2012image, zhang2018ffdnet, ulyanov2018deep}. Deep methods such as autoencoders \cite{lore2017llnet}, multi-layer perceptrons \cite{burger2012image}, and inpainting based denoising autoencoders \cite{xie2012image} have also been investigated. Still, their performance on real-world raw low light image datasets have been sub-par. One interesting direction was investigated in \cite{Brooks_2019_CVPR}, where the noise model was learned from sRGB images, and a subsequently learned denoising pipeline was applied to obtain the enhanced image. While their work explored the learning of relevant photometric parameters in 'unprocessing' and then processing the image, our goal in effectively learning the camera's color and noise models is different. We propose a few-shot domain adaptation method across camera sensors (Bayer filter array) and ISO settings.

\textbf{Few-shot Domain Adaptation}: Few-shot learning \cite{sung2018learning, finn2017model, sun2019meta, prabhakar2021labeled} and Domain adaptation \cite{bousmalis2017unsupervised, rozantsev2018beyond} techniques are well explored in the context of many computer vision tasks. Several few-shot domain adaptation works \cite{sahoo2018meta,motiian2017few}  use few labeled samples with many unlabeled samples in the target domain for image classification. Similarly, DA-FSL \cite{zhao2021domain} is a few-shot domain adaptive prototypical learning method for recognition. The meta-learning paradigm \cite{casas2019few} has also shown great promise in image denoising but depends on prior noise models to partially represent real noise. To the best of our knowledge, there has been no prior investigation of few-shot DA in inverse-imaging for raw camera domains. % Few-shot learning \cite{sung2018learning, finn2017model, sun2019meta} and Domain Adaptation \cite{bousmalis2017unsupervised, rozantsev2018beyond} methods have performed sub-optimally for the joint task of adapting to a poorly represented target with domain-shift in extreme low-light conditions where the noise and distortion are severe.

As noted in \cite{Brooks_2019_CVPR}, different camera sensors exhibit different noise models, and the process of capturing short-exposure and corresponding long-exposure raw images in low-light conditions is expensive and time-consuming. While \cite{wang2020practical} has proposed efficient low-light enhancement, it is only for one type of smartphone camera. As a step toward tackling these challenges, we introduce the first of its kind few-shot domain adaptation and enhancement method for low-light conditions in the raw domain that is lightweight and highly effective.

%  One other caveat pointed out in \cite{jiang2021enlightengan} is that in working with low-light image enhancement in the raw domain, the model learns the pipeline inclusive of color transformations, demosaicing, and denoising and primarily aims to alleviate the artefacts from the long-exposure ground truth images.
        
\vspace{-0.5cm}\section{Proposed method}
\label{sec:method}
    % \begin{table}[t]
% \centering
% \caption{Details of the datasets used in our work.}
% \label{tab:datasets}
% \begin{tabular}{lccc}
% \hline
% Datasets & \begin{tabular}[c]{@{}c@{}}Exposure\\ Ratios\end{tabular} & \begin{tabular}[c]{@{}c@{}}Training\\ images\end{tabular} & \begin{tabular}[c]{@{}c@{}}Validation\\ images\end{tabular} \\ \hline
% Sony \cite{chen2018learning} & 90,150,300 & 161 &  36\\
% Nikon &  100, 300 & 53 &  24 \\
% Canon \cite{CanonLSID} & 50, 150, 300 & 44 &  21\\ \hline
% \end{tabular}
% \end{table}
%--------------------------------------------------------------------------------------------------------------------
% \begin{figure*}
% \begin{center}
% \includegraphics[width=\textwidth, clip, trim=0cm 15.55cm 0.75cm 0.05cm]{figures/FDA-LSID.png}
% \end{center}
%   \caption{Proposed few-shot domain adaptation model architecture.}
% \label{fig:model}
% \end{figure*}
% % \begin{figure*}
% % \begin{center}
% % \includegraphics[scale=0.42]{figures/convert.png}
% % \end{center}
% %   \caption{Source camera specific 16-to-8-bit converter.}
% % \label{fig:convert}
% % \end{figure*}
\begin{figure}[t]
\begin{tabular}{cc}
\includegraphics[page=1, width=5.8cm]{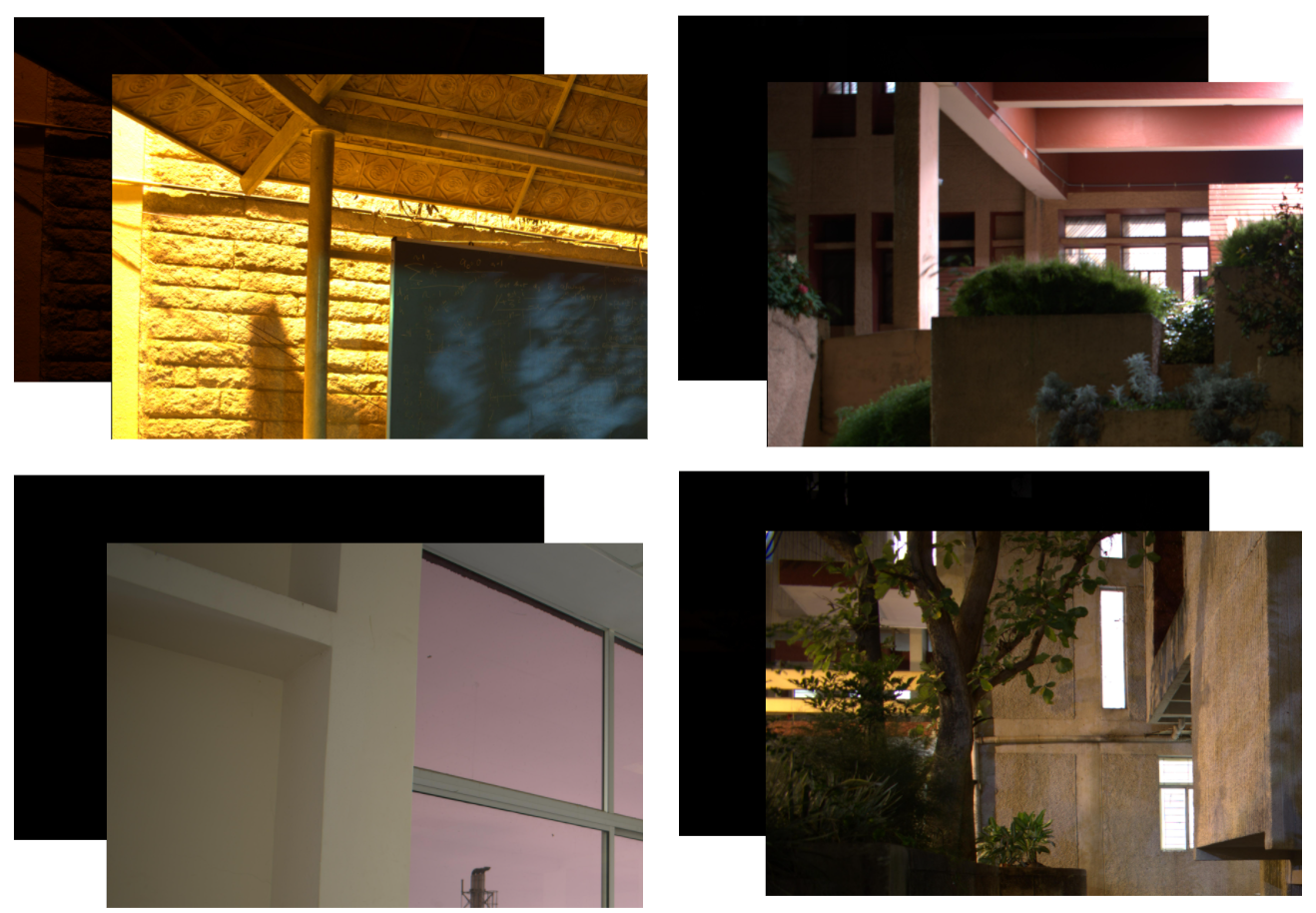}&\hspace{+1mm}
\includegraphics[page=1, clip, trim=0.6cm 16.75cm 7.5cm 0.15cm, scale=0.9]{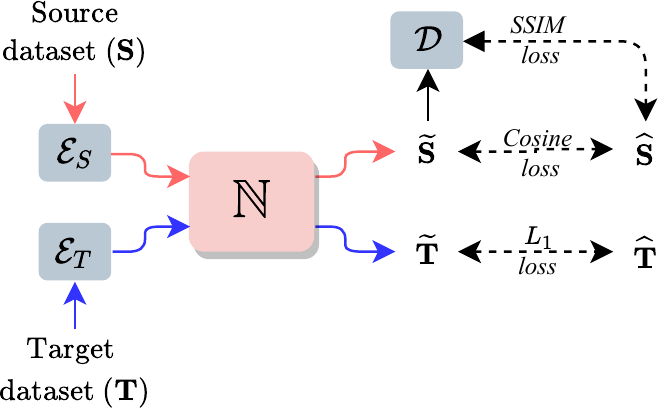}\\
(a)&(b)
\end{tabular}
\caption{(a) Example short-exposure and long-exposure image pairs from the Nikon dataset. The short exposure images are almost entirely dark whereas the long-exposure images have immense scene information. (b) Overview of our few-shot domain adaptation method.}
\label{fig:nikon}
\label{fig:prop_overview}
\end{figure}
With a noisy raw image captured with low-exposure time (i.e., shutter speed) as input, our CNN-based approach is trained to predict a clean long-exposure sRGB output of the same scene. The input is multiplied by an exposure factor calculated by the ratio of output and input exposure times. For example, to generate a 10-second long exposure output, the input 0.1-second low exposure image must be multiplied by 100. As a result of this operation, along with illumination, the noise is also amplified proportionally. Since we multiply the factor in the unprocessed raw domain and expect the output in the sRGB domain, the network must learn camera hardware-specific enhancement as well as its entire ISP pipeline (lens correction, demosaicing, white balancing, color manipulation, tone curve application, color space transform, and Gamma correction). Thus, a model trained on one specific camera data (source domain) does not translate similar performance to a different camera (target domain), hence the domain gap. In this paper, we propose to transfer the enhancement task from large labeled source data and generate output in the target domain using few labeled target data.

\textbf{Problem formulation}: We denote source domain ($\mathbf{S}$) with input short-exposure images as $\{S_n\}$ and corresponding long-exposure ground truth as $\widehat{\mathbf{S}}\!=\!\widehat{S}_n$, $\forall n=1,\cdots,N$. Similarly, the target domain ($\mathbf{T}$) consists of input images $\{T_m\}$ and corresponding ground truth, $\widehat{\mathbf{T}}\!=\!\widehat{T}_m$, $\forall m=1,\cdots,M$. Note that $N$ is much greater than $M$, $N\gg M$. With both $\mathbf{S}$ and $\mathbf{T}$ as input, we train a CNN model ($\mathbb{N}$) to generate enhanced long-exposure output ($\widetilde{\mathbf{S}}$ and $\widetilde{\mathbf{T}}$). Our method is illustrated in Fig. \ref{fig:prop_overview}(b) with the source and target training pipelines. It is an end-to-end trainable deep network that takes the raw sensor arrays as input and performs image enhancement utilizing the source data for few-shot domain adaptation to the target data.

\textbf{Encoders}: \label{sec:pipeline} The significant domain gap between the source and target domains necessitates the extraction of separate and independent features from each domain before processing with a shared enhancement network ($\mathbb{N}$). Hence, we use a source encoder ($\mathcal{E}_S$) and a target encoder ($\mathcal{E}_T$). We first pack the input raw sensor arrays into a four-channel vector (for Bayer arrays from Sony, Nikon, and Canon cameras) and subtract the black level (reference voltage). Then, the packed array is multiplied by the exposure ratio factor and passed as input to the respective domain encoder. It should be noted that the exposure ratio factors need not be the same between the source and the target domain (See Table \ref{tab:datasets}). For the encoder network, we use three convolutional layers with $\{16,32,64\}$ filters and 3$\times$3 kernel size. 

\textbf{Enhancement Network}: The source and target domain encoder features are passed separately to a shared common enhancement network, $\mathbb{N}$. By having a common enhancement network, the large pool of source data helps to improve the enhancement quality of $\mathbb{N}$, while the few target samples ensure that the output is in the target domain. We use U-Net architecture for the enhancement network. Further, the network has a pixel shuffle layer to convert 12-channel prediction to 16-bit three channel sRGB output. The objective of $\mathbb{N}$ is to enhance, denoise, perform other ISP operations (AWB, color manipulation, etc.), and finally demosaicking to generate an sRGB output. $\mathbb{N}$ generates enhanced output $\widetilde{\mathbf{T}}$ for the target domain data as, $\widetilde{\mathbf{T}} = \mathbb{N}\big(\mathcal{E}_T(\mathbf{T}) \big)$. Similarly, $\widetilde{\mathbf{S}}$ for the source domain as, $\widetilde{\mathbf{S}} = \mathbb{N}\big(\mathcal{E}_S(\mathbf{S}) \big)$.
%\footnote{Please refer to the supplementary material for detailed network definition.}

\textbf{Losses}: For the target domain, we compute the $\ell_1$ loss between the prediction ($\widetilde{\mathbf{T}}$) and the ground truth ($\widehat{\mathbf{T}}$) as, $\mathcal{L}_{target} = \ell_1\big(\widetilde{\mathbf{T}},\widehat{\mathbf{T}} \big)$. The source domain loss consists of two components: cosine similarity loss and SSIM loss. We compute cosine similarity between $\widetilde{\mathbf{S}}$ and $\widehat{\mathbf{S}}$ as, 
$
    \mathcal{L}_{CS}(\widetilde{\mathbf{S}},\widehat{\mathbf{S}})= 1 -  \frac{{\widetilde{\mathbf{S}} \cdotp \widehat{\mathbf{S}}}}{\|\widetilde{\mathbf{S}}\|\times\|\widehat{\mathbf{S}}\|}
$.
% \begin{equation}
%     \mathcal{L}_{CS}(\widetilde{\mathbf{S}},\widehat{\mathbf{S}})= 1 -  \frac{{\widetilde{\mathbf{S}} \cdotp \widehat{\mathbf{S}}}}{\|\widetilde{\mathbf{S}}\|\times\|\widehat{\mathbf{S}}\|}
% \end{equation}
Cosine similarity loss is weak supervision for the source domain and is used instead of $\ell_1$ loss since $N\gg M$, and using a strong supervision loss like $\ell_1$ optimizes for pixel values to train $\mathbb{N}$, making the network predict the output in the source domain even for target domain input. Cosine similarity loss ensures that the prediction and the ground truth are in a similar direction. Hence with $\mathcal{L}_{CS}$, $\mathbb{N}$ can still perform enhancement while predicting in target domain even for source domain input. Further, when trained with Sony as source and 4-shot Nikon as target (Table \ref{tab:ablation}) with $L_1$ loss for the source, we obtain only 27.14dB PSNR for target domain validation, whereas using $\mathcal{L}_{CS}$ loss for source achieves 30.30dB PSNR.

% \begin{figure*}[t]
%     \centering
%     \includegraphics[width=\linewidth]{3_BMVC/Images/Canon-Results-1.pdf}
%     \caption{Qualitative comparison with methods tested on Canon target images. (b)HDRCNN and (c) Unprocess are trained on Sony source and fine-tuned on 6-Canon target images, LSID with (d) 6-Canon target images and (e) full ($k$=44) Canon training dataset, (f) Proposed few-shot domain adaptation approach with 6-Canon target images and 161 Sony source images.}
%     \label{fig:canon_eg2}
% \end{figure*}

\begin{figure*}[t]
\centering
\subfigure{\includegraphics[width=\linewidth]{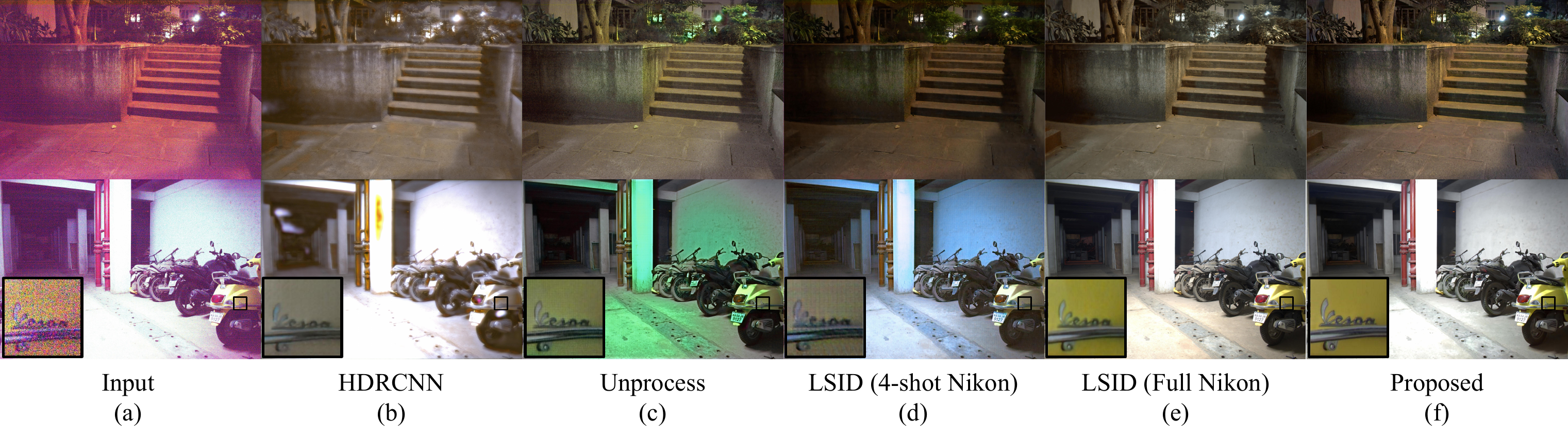}}\\ \vspace{-1.65\baselineskip}
\subfigure{\includegraphics[width=\linewidth]{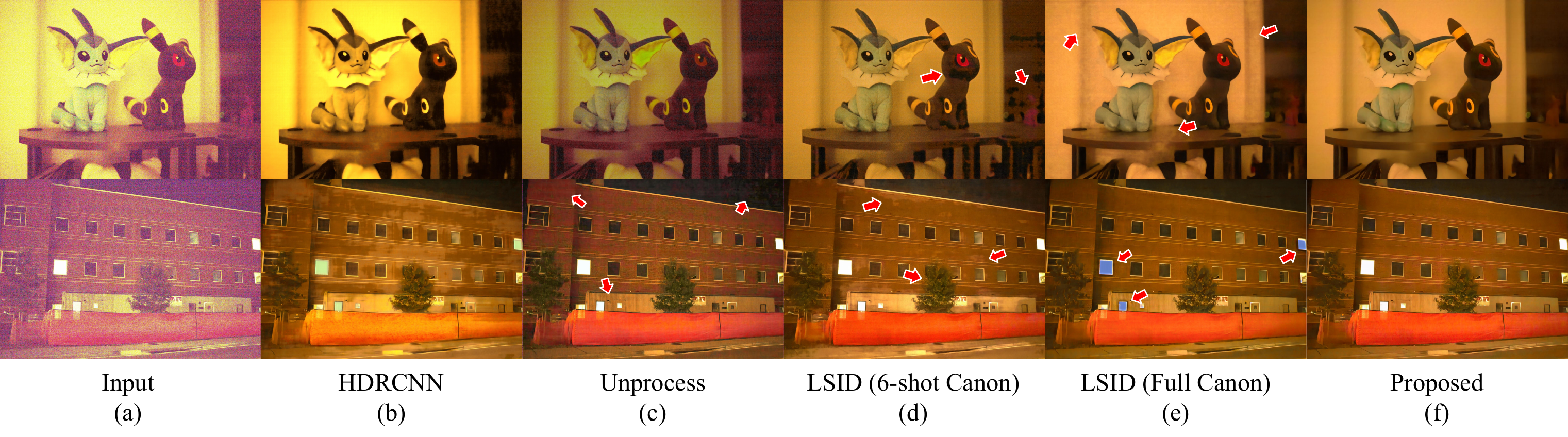}}
\caption{Qualitative comparison with methods tested on Nikon (top row) and Canon (bottom row) target images. (a) Input after multiplying by exposure factor, results from (b) HDRCNN and (c) Unprocess methods are after training on full Sony source and fine-tuning on $k$-shot target images, LSID with (d) $k$-shot target images and (e) full target training dataset ($k$=53 for Nikon and $k$=44 for Canon), (f) Proposed few-shot domain adaptation method with 161 Sony source images and 4-shot Nikon (top row) and 6-Canon (bottom row) target images.}
\label{fig:nikon_eg1}
\label{fig:canon_eg2}
\end{figure*}

From experiments (in section \ref{sec:exp}), we find better enhancement (in terms of PSNR) using the structural similarity index measure (SSIM) \cite{wang2004image} to compute perceived degradation and preserve the spatial structure in the source output with respect to the ground truth. We do not use SSIM directly on the 16-bit data as that causes the source data to heavily influence the domain adaptation since the source dataset is much larger. Hence, we apply SSIM in JPEG compressed 8-bit domain, where the structural domain difference is less. Since type-casting the 16-bit data to 8-bit will still possess domain-specific details, we train a 16-to-8-bit U-net model ($\mathcal{D}$ in Fig. \ref{fig:prop_overview}) to convert the output from 16-bit to post-processed 8-bit representation. 

The $\mathcal{D}$ network is trained to perform the following non-linear operations: White balancing, Gamma correction, Quantization, and JPEG compression. Even after JPEG compression, the prediction may have traces of source domain specific color information. Further, the SSIM loss is a strong pixel-wise supervision, and in order to avoid the source domain from heavily influencing $\mathbb{N}$, we compute SSIM loss only in grayscale space, not in RGB color space. Also, it follows the intuition that the structure and edge information of a scene will remain the same across images captured with different cameras, while the color space representation may vary. We find that without SSIM loss for the source, we obtain 29.38dB PSNR on target domain validation, whereas using SSIM loss achieves 30.30dB PSNR (Table \ref{tab:ablation}). For computing the SSIM loss, the ground truth ($\widehat{\mathbf{S}}$) is also converted offline to post-processed 8-bit data ($\widehat{\mathbf{S}}_{PP}$) using the rawpy post process function. Hence, the loss is obtained by computing SSIM loss between $\mathcal{D}(\widetilde{\mathbf{S}})$ and $\widehat{\mathbf{S}}_{PP}$,
$
    \mathcal{L}_{SSIM} = 1 - SSIM\Big(\mathcal{D}(\widetilde{\mathbf{S}}), \widehat{\mathbf{S}}_{PP}\Big)
$.
% \begin{equation}
%     \mathcal{L}_{SSIM} = 1 - SSIM\Big(\mathcal{D}(\widetilde{\mathbf{S}}), \widehat{\mathbf{S}}_{PP}\Big)
% \end{equation}
In Fig. \ref{fig:prop_overview}, the top branch guided by the deep red arrows shows the entire source camera training pipeline. It should be noted that $\mathcal{D}$ is used only to compute the loss but not in inference. Finally, we use the sum of cosine similarity loss ($\mathcal{L}_{CS}$) as well as the SSIM loss calculated in the 8-bit domain as the total loss for the source camera pipeline: $\mathcal{L}_{source} = \mathcal{L}_{CS} + \mathcal{L}_{SSIM}$. The total loss is the sum of target and source domain losses: $\mathcal{L}_{total}=\mathcal{L}_{target}+\mathcal{L}_{source}$.

\vspace{-0.5cm}\section{Experiments}
\label{sec:data}
    % \subsection{Source and target datasets} 
\label{section:fsc}
\textbf{Datasets}: We use the Sony camera dataset \cite{chen2018learning} for our source training pipeline. We expect the diverse, high-quality low-light scenes from this dataset to aid few-shot domain adaptation performance in terms of the color spaces and noise model learned by our method. 
\begin{table}[t]
\centering
\caption{Quantitative comparison of Sony as source with Nikon or Canon as target. The improvement of our method over only $k$ shot trained model is in brackets. The LSID model trained with full Nikon dataset ($k$=53) achieves 30.74dB PSNR and 0.803 SSIM and when trained with full Canon dataset ($k$=44) attains 32.32dB PSNR and 0.899 SSIM (see Table \ref{tab:baseline}).}
\label{tab:nikon}
\label{tab:canon}
\scalebox{0.735}{
\begin{tabular}{@{}lccc|ccc@{}}
\thickhline
\textbf{Nikon as target} & \multicolumn{3}{c|}{PSNR} & \multicolumn{3}{c}{SSIM} \\ \hline
$k$ ($\rightarrow$) & 1 & 2 & 4 & 1 & 2 & 4 \\ \hline
\begin{tabular}[c]{@{}l@{}}LSID\\ (only $k$ target)\end{tabular} & 23.20 $\pm$ 3.06 & 27.27 $\pm$ 0.384 & 28.05 $\pm$ 1.53 & 0.679 $\pm$ 0.172 & 0.819 $\pm$ 0.031 & 0.864 $\pm$ 0.011 \\ \hline
\begin{tabular}[c]{@{}l@{}}Proposed\\ ($k$ target + source)\end{tabular} & \begin{tabular}[c]{@{}c@{}}\textbf{25.27} $\pm$ 0.58\\ (+2.07)\end{tabular} &
\begin{tabular}[c]{@{}c@{}}\textbf{28.06} $\pm$ 0.671\\ (+0.79)\end{tabular} &
\begin{tabular}[c]{@{}c@{}}\textbf{30.30} $\pm$ 0.52\\ (+2.25)\end{tabular} & \begin{tabular}[c]{@{}c@{}}\textbf{0.860} $\pm$ 0.010\\ (+0.181)\end{tabular} &
\begin{tabular}[c]{@{}c@{}}\textbf{0.909} $\pm$ 0.003\\ (+0.090)\end{tabular} &
% \begin{tabular}[c]{@{}c@{}}\textbf{0.913} $\pm$ 0.006\\ (+0.049)\end{tabular} \\ \midrule
% \begin{tabular}[c]{@{}l@{}}LSID\\ (full target, $k$ = 53)\end{tabular} & \multicolumn{3}{c|}{30.74} & \multicolumn{3}{c}{0.803} \\ \bottomrule
\begin{tabular}[c]{@{}c@{}}\textbf{0.913} $\pm$ 0.006\\ (+0.049)\end{tabular} \\ \hline \hline
\textbf{Canon as target} &  &  &  &  &  &  \\ \hline
$k$ ($\rightarrow$) & 1 & 3 & 6 & 1 & 3 & 6 \\ \hline
\begin{tabular}[c]{@{}l@{}}LSID\\ (only $k$ target)\end{tabular} & 21.54 $\pm$ 2.89 & 26.9 $\pm$ 2.37 & 29.36 $\pm$ 0.763 & 0.588 $\pm$ 0.182 & 0.785 $\pm$ 0.005 & 0.829 $\pm$ 0.007 \\ \hline
% \begin{tabular}[c]{@{}l@{}}Proposed\\ ($k$ target + source)\end{tabular} & \begin{tabular}[c]{@{}c@{}}\textbf{24.29} $\pm$ 3.16\\ (+2.75)\end{tabular} & \begin{tabular}[c]{@{}c@{}}\textbf{28.78} $\pm$ 3.54\\ (+1.8)\end{tabular} & \begin{tabular}[c]{@{}c@{}}\textbf{33.22} $\pm$ 0.45\\ (+3.86)\end{tabular} & \begin{tabular}[c]{@{}c@{}}\textbf{0.623} $\pm$ 0.0074\\ (+0.035)\end{tabular} & \begin{tabular}[c]{@{}c@{}}\textbf{0.841} $\pm$ 0.0335\\ (+0.056)\end{tabular} & \begin{tabular}[c]{@{}c@{}}\textbf{0.896} $\pm$ 0.015\\ (+0.067)\end{tabular} \\ \midrule
% \begin{tabular}[c]{@{}l@{}}LSID\\ (full target, $k$ = 45)\end{tabular} & \multicolumn{3}{c|}{32.32} & \multicolumn{3}{c}{0.899} \\ \bottomrule
\begin{tabular}[c]{@{}l@{}}Proposed\\ ($k$ target + source)\end{tabular} & \begin{tabular}[c]{@{}c@{}}\textbf{24.29} $\pm$ 3.16\\ (+2.75)\end{tabular} & \begin{tabular}[c]{@{}c@{}}\textbf{28.78} $\pm$ 3.54\\ (+1.8)\end{tabular} & \begin{tabular}[c]{@{}c@{}}\textbf{33.22} $\pm$ 0.45\\ (+3.86)\end{tabular} & \begin{tabular}[c]{@{}c@{}}\textbf{0.623} $\pm$ 0.007\\ (+0.035)\end{tabular} & \begin{tabular}[c]{@{}c@{}}\textbf{0.841} $\pm$ 0.033\\ (+0.056)\end{tabular} & \begin{tabular}[c]{@{}c@{}}\textbf{0.896} $\pm$ 0.015\\ (+0.067)\end{tabular} \\ \thickhline
\end{tabular} }
\end{table}

% \begin{table}
% \begin{center}
% \begin{tabular}{|l|c|c|c|}
% \hline
% Camera & Array & \#Train & \#Test & Ratios \\
% \hline\hline
% Sony & Bayer & 161 & 37 & 100\\
% Sony & Bayer & 161 & 37 & 100, 250, 300\\
% Sony & Bayer & 161 & 37 & 100, 250, 300\\
% \hline
% \end{tabular}
% \end{center}
% \caption{Dataset Info. }
% \end{table}

% \begin{figure*}
% \begin{center}
% \includegraphics[scale=0.4]{figures/fuji_nikon.png}
% \includegraphics[scale=0.4]{figures/fuji_nikon.png}
% \end{center}
%   \caption{Proposed few-shot domain adaptation model architecture.}
% \label{fig:model}
% \end{figure*}

%---------------------------------------------------------------------------------------------------------------
% \begin{figure}[t]
% \begin{center}
% \includegraphics[width=0.7\linewidth]{3_BMVC/Images/nikon.png}
% \end{center}
%   \caption{Examples of short-exposure and long-exposure image pairs from the Nikon dataset. Note that the short exposure images are almost entirely dark whereas the long-exposure images possess immense scene-related information.}
% \label{fig:nikon}
% \end{figure}

For few-shot domain adaptation, we work with very few (< 10) target camera images in every training experiment. We use the open-source Canon camera low-light raw image dataset \cite{CanonLSID} and a new Nikon camera dataset that we have compiled and make available with this work for our target camera training pipeline. We do not investigate burst denoising or the `lucky imaging' phenomenon. Hence, we only take the first short-exposure raw image for each scene from the Sony dataset and use the 161 images for our source camera training pipeline. Note that the Canon dataset has eight different ratios with close ranges such that they can be put into three buckets of ratios: 50, 150, and 300 (Table \ref{tab:datasets}). % We also show qualitative results for low-cost smartphone camera sensors - Google Pixel and OnePlus 5. 
%, to evaluate our approach on severe noise from low-cost smartphone camera sensors. %We also modify the Canon camera dataset to increase the number of training images by transferring a few validation images to the train set. This allows our baselines to learn the camera parameters in a better manner while we stick to few-shot training. 

% We also experiment with raw low-light images taken with smartphone cameras - Google Pixel and OnePlus 5, to investigate the performance of our proposed approach on raw images captured with low-cost smartphone camera sensors, which typically have higher noise severity in low-light conditions.

\textbf{Nikon camera dataset}:
\label{section:nikon}
We have compiled a dataset of raw low-light images captured with a Nikon D5600 camera to train the proposed few-shot domain adaptation architecture. The Nikon dataset consists of short-exposure images captured at $\nicefrac{1}{3}$ or $\nicefrac{1}{10}$ seconds and corresponding ground-truth long-exposure images captured at 10 or 30 seconds in the NEF format. For uniformity, there are two short-exposure images for every long-exposure image such that the exposure ratio (ratio of exposure time between the ground-truth long-exposure image and the input short-exposure image) is 100 and 300, respectively. Similar to \cite{chen2018learning}, we mount the camera on sturdy tripods and use appropriate camera settings to capture the static scenes using a smartphone app. The images captured include 129 short-exposure and 65 long-exposure ground-truth images of indoor and outdoor low-light scenes (sub lux). %Our aim in compiling the Nikon camera raw image dataset is to capture realistic low-light scenes and investigate few-shot domain adaptation with a dataset containing few images. %As discussed in section \ref{sec:related}, several related works compare results with datasets containing as few as only up to 10 images, and we seek to replicate these experimental setups. 

\textbf{Training Setup:} \label{sec:train}
% \textbf{Training $\mathcal{D}$}: The 16-to-8 bit converter used for the source pipeline is a fully-convolutional U-net \cite{ronneberger2015u} network architecture, trained with the ground-truth image pairs of 16-bit and 8-bit representations. We use the $L_{2}$ loss for training the converter and train it for 4000 epochs. We use the trained 16-to-8-bit converter for our model training.
We train the source and target pipelines simultaneously in an end-to-end manner. As discussed in section \ref{sec:pipeline}, we use the respective short-exposure raw images as the input to each of the encoders. We first randomly crop a $512\times512$ image patch and augment it with random-flip and random-rotate. We then subtract the black level and multiply the input raw image with the exposure ratio. We use an initial learning rate of $10^{-4}$ up to 2000 epochs and then reduce it by a factor of 10 for every 1000 epochs thereafter. We use the Adam \cite{kingma2014adam} optimizer for the 8-bit SSIM loss and Cosine Similarity loss ($\mathcal{L}_{source}$) for the source pipeline, and the $\mathcal{L}_{target}$ loss for the target pipeline.

We train the model for 4000 epochs (same as \cite{chen2018learning}) but observe the loss saturating at lower epochs prompting us to employ early stopping. Since the large source domain has 161 images, every epoch has 161 train steps. As we jointly train the source and target pipelines, for every epoch, we use 161 randomly cropped source patches obtained from the 161 source domain raw images and 161 randomly cropped target patches from only $k$-images in the target domain. We find that training our proposed method for up to 2500 epochs is sufficient to obtain the best results and reproduce the results in this paper for few-shot domain adaptation.

The source SSIM loss is calculated in the 8-bit space after passing the output from the shared $\mathbb{N}$ through the 16-to-8-bit converter. The cosine similarity loss for the source domain and $L_{1}$ loss for the target domain are computed in the 16-bit sRGB space. The exposure ratio is computed and provided to the network. At inference time, we use the full-scale raw target image as input to the target camera pipeline and obtain the enhanced target sRGB image. % in the 16-bit sRGB space.
% \begin{table}[t]
%     \centering
%     \caption{Details of the datasets used in our work.}
%     \label{tab:datasets}
%     \scalebox{0.85}{
%     \begin{tabular}{lccc}
%     \hline
%     \multirow{2}{*}{Datasets} & Exposure & Training & Validation \\
%     & Ratios & Images & Images \\
%     \hline
%     Sony \cite{chen2018learning} & 90,15,300 & 161 & 36\\
%     Nikon & 100,300 & 53 & 24\\
%     Canon \cite{CanonLSID} & 50,150,300 & 44 & 21 \\ \hline
%     \end{tabular}
%     }
%     \vspace{-2.5mm}
% \end{table}

% The images in our proposed dataset was captured with ISO 100-200. Other datasets were captured with ISO between 50-200. The range of ISO does not affect the model's performance due to the wide range of amplification factors used to train the model. A single encoder is sufficient to handle all ISO values.

% \begin{table}[t]
% \centering
% \begin{tabular}{lccc}
% \hline
% Datasets & \begin{tabular}[c]{c}Exposure\\ Ratios\end{tabular} & \begin{tabular}[c]{c}Training\\ images\end{tabular} & \begin{tabular}[c]{c}Validation\\ images\end{tabular} \\ \hline
% Sony & 90,150,300 & 161 &  36\\ %\cite{chen2018learning}
% Nikon &  100, 300 & 53 &  24 \\
% Canon & 50, 150, 300 & 44 &  21\\ \hline %\cite{CanonLSID}
% \end{tabular}
% \caption{Details of the datasets used in our work.}
% \label{tab:datasets}
% \end{table}
    
\vspace{-0.5cm}\section{Results}
\label{sec:exp}
    % \textbf{I would like to see a motivation why the models trained on the full Canon and
% Nikon datasets underperform the models trained with a domain transfer (e.g.
% Figure5). This might suggests the proposed algorithm is indeed doing a great
% job, but since it is counter-intuitive, it deserves an expanded discussion.}
% \usepackage{booktabs}
\begin{figure*}[t]
\centering
\subfigure{\includegraphics[width=\linewidth]{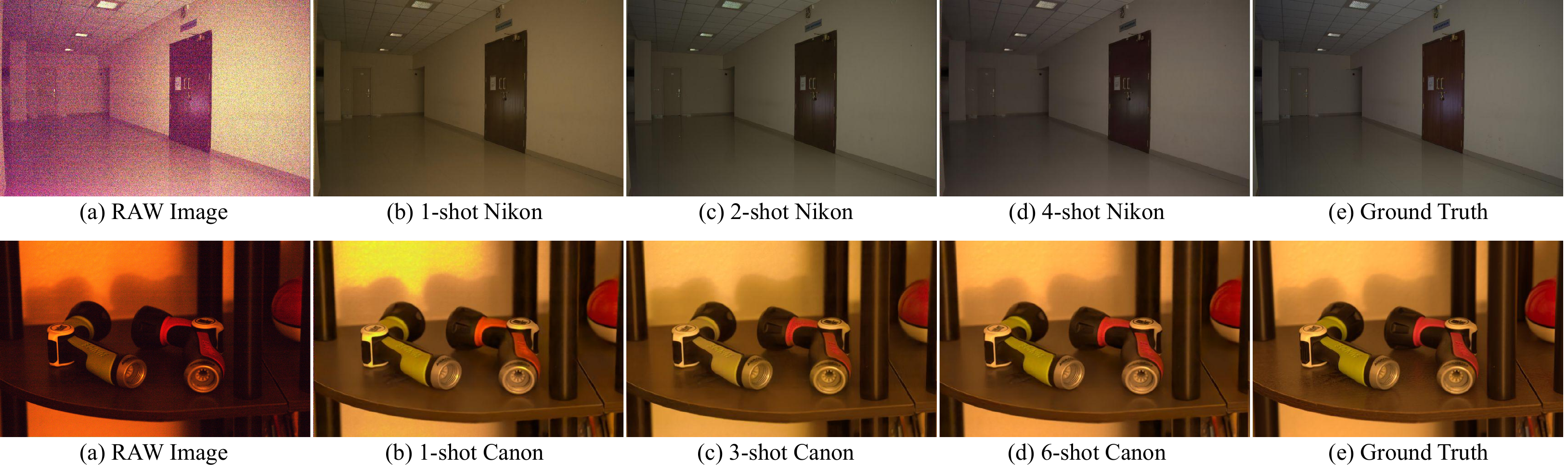}}
\caption{Qualitative comparison of results for $k$=1,2 and 4 for Sony as source and Nikon as target, and $k$=1,3 and 6 for Sony as source and Canon as target for choosing the value of $k$.}
\label{fig:canon136}
\end{figure*}

\begin{table}[t]
\centering
\caption{Comparison with baselines trained with Sony raw data as source and fine-tuned with (1/2/4-shot) Nikon camera raw images and (1/3/6-shot) Canon camera raw images. We compare with HDRCNN \cite{EKDMU17}, Unprocess \cite{Brooks_2019_CVPR} and LSID \cite{chen2018learning}.}
\label{tab:finetunenikon}
\label{tab:finetunecanon}
\scalebox{0.71}{
\begin{tabular}{c|c}
    \begin{tabular}{|c|cccccc}
        \hline
        Method & \multicolumn{6}{c}{Sony source w/ Nikon target} \\ \hline
        \multirow{1}{*}{$k$ ($\rightarrow$)} & \multicolumn{2}{c}{1} & \multicolumn{2}{c}{2} & \multicolumn{2}{c}{4} \\
        \hline
        & PSNR & SSIM & PSNR & SSIM & PSNR & SSIM  \\ \hline
        HDRCNN & 11.47 & 0.648 & 12.36 & 0.578 & 12.87 & 0.627 \\ 
        Unprocess & 18.90 & 0.710 & 21.91 & 0.690 & 25.63 & 0.761 \\ 
        LSID & 22.38 & 0.746 & 25.75 & 0.874 & 27.93 & 0.899 \\ 
        Proposed & \textbf{25.27} & \textbf{0.860} & \textbf{28.06} & \textbf{0.909} & \textbf{30.30} & \textbf{0.913} \\ 
        \hline
    \end{tabular}
    &
    \begin{tabular}{cccccc|}
        \hline
        \multicolumn{6}{c|}{Sony source w/ Canon target} \\ \hline
        \multicolumn{2}{c}{1} & \multicolumn{2}{c}{3} & \multicolumn{2}{c|}{6} \\ 
        \hline
        PSNR & SSIM & PSNR & SSIM & PSNR & SSIM  \\ \hline
        15.07 & 0.593 & 15.89 & 0.627 & 16.14 & 0.633 \\ 
        16.97 & 0.609 & 22.15 & 0.671 & 27.20 & 0.733\\ 
        \textbf{25.88} & \textbf{0.792} & 28.38 & 0.831 & 28.85 & 0.819 \\ 
        24.29 & 0.623 & \textbf{28.78} & \textbf{0.841} & \textbf{33.22} & \textbf{0.896} \\
        \hline
    \end{tabular}
\end{tabular}
}
\end{table}

\begin{table}[t]
\centering
\caption{Comparison with recent low-light image enhancement methods. Ours is the first few-shot domain adaptation method, hence, we first train the baselines on the Sony camera dataset and then fine-tune with (4-shot) Nikon camera or (6-shot) Canon camera datasets.}
\label{tab:recent}
\scalebox{0.8}{
\begin{tabular}{@{}lccccccc@{}}
\hline
\multirow{1}{*}{Method} & \multicolumn{2}{c}{Sony Source} & \multicolumn{2}{c}{Sony w/ Nikon} & \multicolumn{2}{c}{Sony w/ Canon} \\ \hline
& PSNR & SSIM & PSNR & SSIM & PSNR & SSIM \\
%  & & & Canon & Nikon \\
%  & & & Target & Target \\
\hline
%  & HDRCNN \cite{EKDMU17} &  12.64 & 11.78 & \\
DeepUPE \cite{Wang_2019_CVPR} & 14.58 & 0.256 & 13.42 & 0.266 & 13.81 & 0.285\\
MIRNet \cite{Zamir2020MIRNet} & 15.24 & 0.414 & 14.18 & 0.458 & 13.24 & 0.397 \\
KnD \cite{zhang2019kindling} & 17.15 & 0.313 & 15.04 & 0.226 & 17.24 & 0.432 \\
HDRCNN \cite{EKDMU17} & 17.39 & 0.491 & 12.87 & 0.627 & 16.14 & 0.633 \\
KnD++ \cite{zhang2021beyond} & 23.03 & 0.579 & 19.38 & 0.471 & 21.25 & 0.434 \\
Unprocess \cite{Brooks_2019_CVPR} &  27.83 & 0.700 & 25.63 & 0.761 & 27.20 & 0.733 \\
LSID \cite{chen2018learning} &  28.50 & 0.774 & 27.93 & 0.899 & 29.36 & 0.829 \\
% & DeepUPE \cite{Wang_2019_CVPR} &  29.13 & & \\
Proposed &  - & - & \textbf{30.30} & \textbf{0.913} & \textbf{33.22} & \textbf{0.896} \\ \hline
\end{tabular}
}
\end{table}

Table \ref{tab:datasets} lists the total number of labeled raw image pairs in the train set to be 161 images for the Sony dataset \cite{chen2018learning}, 53 images for our Nikon dataset, and 44 images for the Canon dataset \cite{CanonLSID}. We train our model for three different numbers of labeled target data: For Nikon target, we use $k$=1,2, and 4. For Canon target, we use $k$=1,3, and 6. We choose the $k$ based on the different exposure ratios available for the dataset. As reported in Table \ref{tab:datasets}, the Nikon camera dataset has two exposure ratios; hence we use two images per ratio leading to a total of four images. Similarly, the Canon camera dataset has three ratios; hence we use two images per ratio leading to a total of six images. We observe that using two images per exposure ratio in the target domain is sufficient to outperform all baselines (Refer Table \ref{tab:nikon}). A qualitative guideline for choosing the value of $k$ is in Fig.(\ref{fig:canon136}). For each $k$, we run three separate experiments, each with a different set of $k$ labeled target images. We report the average and 95\% variance margin computed across three different sets for the Nikon dataset and for the Canon dataset (Table \ref{tab:canon}). We compare with an LSID model trained with only $k$-target data and an LSID model trained with the full target camera data. % for a fair comparison.

\textbf{Quantitative Evaluation:} For Nikon dataset as target, our 4-shot approach achieves 30.30dB PSNR, which is on par with the full target dataset ($k$=53) trained LSID model. Our method outperforms only $k$-shot trained model by 2.25dB PSNR (Table \ref{tab:nikon}). Similarly, for Canon dataset as target, our 6-shot approach outperforms the full dataset ($k$=44) trained LSID model by 0.9dB PSNR, and $k$-shot trained LSID model by 3.86dB PSNR (Table \ref{tab:canon}). Since ours is the first few-shot domain adaptation method, we quantitatively compare our method with baselines by training them on the full Sony source dataset and then fine-tuning them in a few-shot manner on the Nikon or Canon target datasets (Table \ref{tab:finetunecanon}). We also compare with recent low-light enhancement methods in Table \ref{tab:recent} and show that our method outperforms all $k$-shot fine-tuned baselines in PSNR and SSIM for the Nikon and Canon target datasets. We show quantitative results for our method trained with Sony as source and four OnePlus camera images or four Google Pixel camera images as target in Table \ref{tab:oneplus}.

As discussed in section \ref{sec:intro}, capturing a low-light raw image dataset is difficult, and different cameras have different color-space and noise distributions, hence there is a need for a domain adaptation method that can tranfer the task from source to a target domain in a few-shot setting. Despite the high complexity of the task, our method outperforms all baselines with a lightweight model because the abundant source data helps to learn the low-light enhancement task in the source+$k$-shot setting successfully as compared to using only target domain data. While the $k$-shot samples help to predict the output in the target domain, the task is transferred successfully from the large source domain to the target camera domain.

\begin{figure}
\begin{tabular}{cc}
\includegraphics[page=1, width=6.15cm, clip, trim=0cm 3.5cm 0cm 0cm]{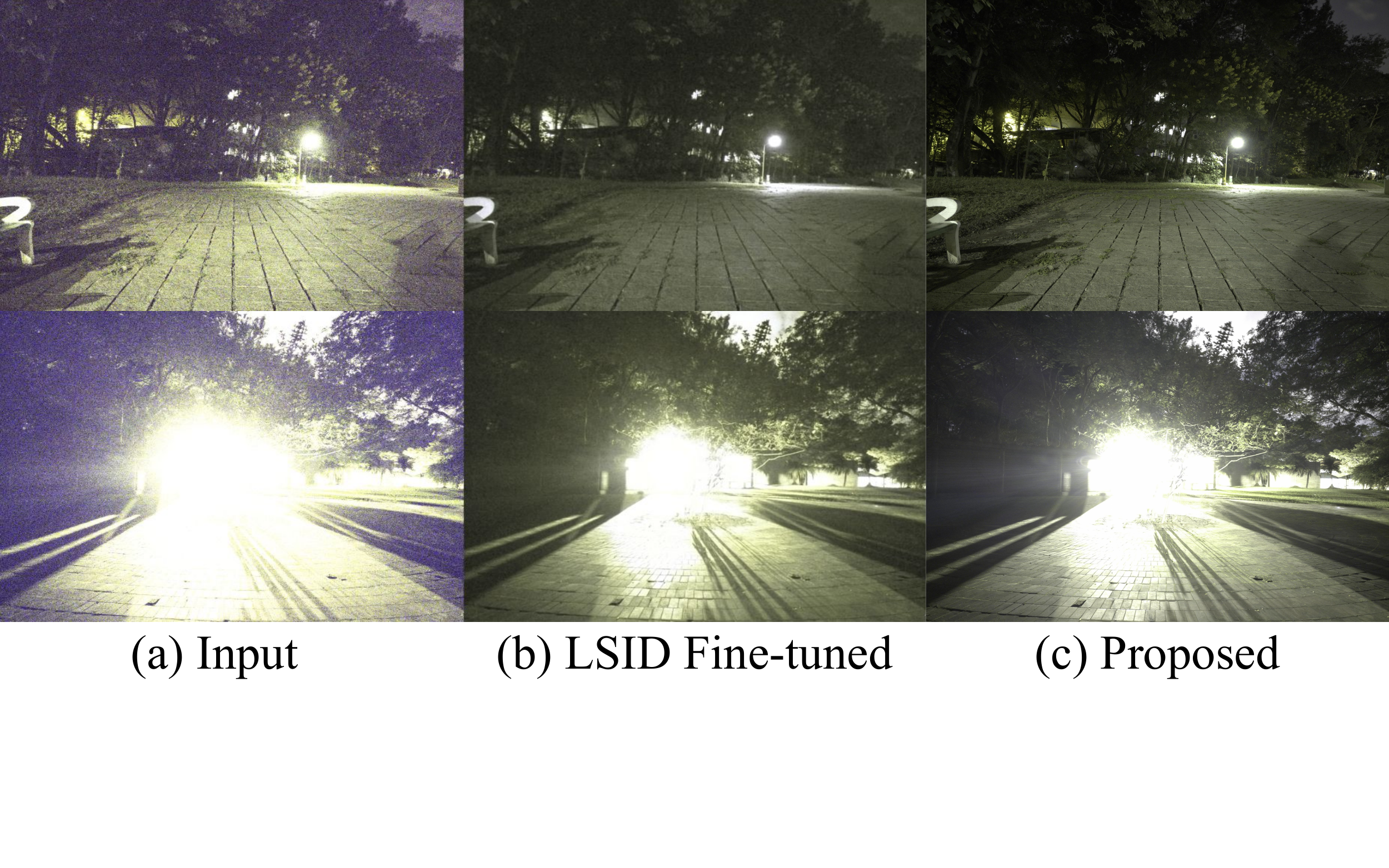}&\hspace{-3.5mm}
\includegraphics[page=1, width=6.15cm, clip, trim=0cm 3.5cm 0cm 0cm]{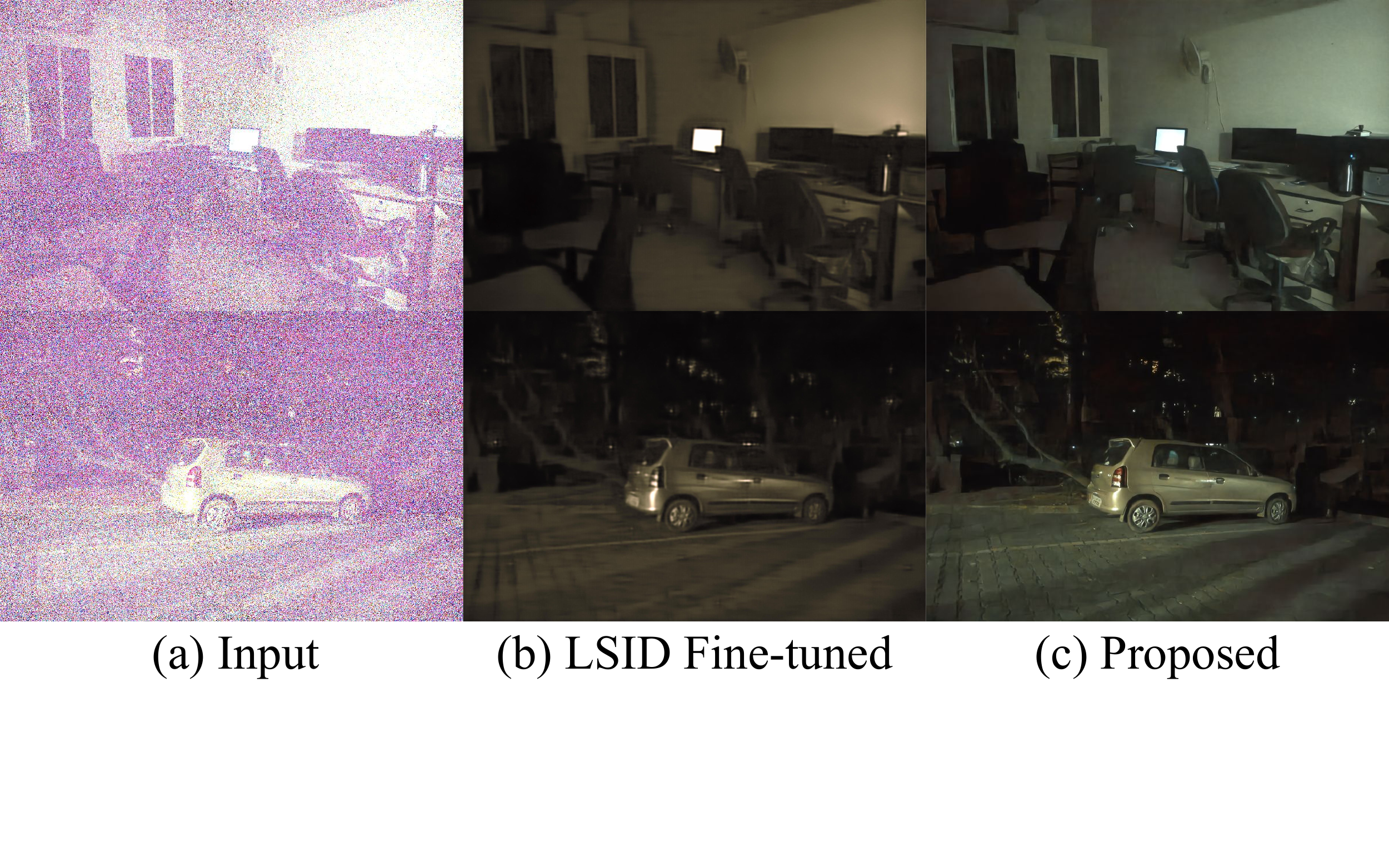}\\
(a)&(b)
\end{tabular}
\caption{Results for Sony as source with (a) OnePlus 5 as target (b) Google Pixel as target.}
\label{fig:oneplus}
\label{fig:pixel}
\vspace{-0.5cm}
\end{figure}

\textbf{Qualitative Evaluation:} \label{subsec:qual} In Fig. \ref{fig:nikon_eg1}, we qualitatively compare the results from our method and baselines when trained with Sony as source and Nikon (top row) or Canon (bottom row) as target. As highlighted by the zoomed-in regions and red arrows, the baseline results have several artifacts in terms of noise and color. Although the LSID model trained with full target data performs better than the $k$-shot model (for Nikon or Canon as target), it is still sub-par compared to our method's results. We also show qualitative results for our method trained with Sony as source and four OnePlus camera images (Fig. \ref{fig:oneplus}(a)) or four Google Pixel camera images (Fig. \ref{fig:pixel}(b)) as target to demonstrate the effectiveness of our method on low-cost smartphone camera data, which typically have higher noise severity in low-light. 
% HDRCNN \cite{EKDMU17}, Unprocess \cite{Brooks_2019_CVPR}, $k$-shot LSID, and full-dataset LSID methods

\begin{table}[t]
\centering
\caption{Quantitative evaluation of Sony as source with OnePlus or Pixel as target. The LSID model trained with full OnePlus dataset ($k$=12) achieves 28.88dB PSNR and 0.708 SSIM and when trained with full Pixel dataset ($k$=10) attains 27.95dB PSNR and 0.759 SSIM. (Note: test set for OnePlus=38 images and Pixel=18 images).}
\label{tab:oneplus}
\label{tab:pixel}
\scalebox{0.71}{
\begin{tabular}{c|c}
    \begin{tabular}{|c|cccccc}
        \hline
        Method & \multicolumn{6}{c}{Sony source w/ OnePlus target} \\ \hline
        \multirow{1}{*}{$k$ ($\rightarrow$)} & \multicolumn{2}{c}{1} & \multicolumn{2}{c}{2} & \multicolumn{2}{c}{4} \\
        \hline
        & PSNR & SSIM & PSNR & SSIM & PSNR & SSIM  \\ \hline
        LSID & 22.54 & 0.617 & 25.87 & 0.651 & 26.69 & 0.670 \\ 
        Proposed & \textbf{25.10} & \textbf{0.633} & \textbf{26.47} & \textbf{0.650} & \textbf{27.77} & \textbf{0.712} \\ 
        % LSID (Full) & \multirow{6}{*}{28.89  0.708} \\
        \hline
    \end{tabular}
    &
   \begin{tabular}{cccccc|}
        \hline
        \multicolumn{6}{c|}{Sony source w/ Pixel target} \\ \hline
        \multicolumn{2}{c}{1} & \multicolumn{2}{c}{2} & \multicolumn{2}{c|}{4} \\
        \hline
        PSNR & SSIM & PSNR & SSIM & PSNR & SSIM  \\ \hline
        13.86 & 0.296 & 21.91 & 0.695 & 27.62 & 0.760 \\ 
        \textbf{18.60} & \textbf{0.635} & \textbf{24.98} & \textbf{0.753} & \textbf{29.60} & \textbf{0.782} \\ 
        % LSID (Full) & \multirow{6}{*}{28.89  0.708} \\
        \hline
    \end{tabular}
\end{tabular}
}
\end{table}

\begin{table}[t]
\centering
% 1 & Finetuning only last layers of LSID &  27.30 $\pm$0.60\\
% 2 & Finetuning all layers of LSID &  27.93 $\pm$0.66\\
\caption{Ablation study for training with Sony source and 4-shot Nikon as target dataset. The table reports mean PSNR/SSIM over three different runs. See section \ref{sec:ablation} for details.}
\label{tab:ablation}
\scalebox{0.85}{
\begin{tabular}{llcc}
\hline
No. & \multicolumn{1}{c}{Details} & \multicolumn{1}{c}{PSNR} & \multicolumn{1}{c}{SSIM} \\ \hline
% 1 & LSID trained w/ full Sony &  28.50 & 0.774 \\
% 2 & CIE-XYZ Input & 27.16 & 0.894 \\
% 3 & Finetuning last layer of LSID &  27.30 & 0.886\\
% 4 & Finetuning all layers of LSID &  27.93 & 0.899\\
1 & Separate encoder and decoder &  28.62 & 0.867\\
2 & Combined encoder & 29.20 & 0.890 \\
% 7 & LAB Input &  27.90 & 0.827 \\
% 5 & With discriminator loss &  28.64\\
% 6 & Pseudo-labeling &  29.44 & 0.907\\
% 3 & 8-bit RGB input & 27.88 & 0.812 \\
% 4 & LAB input &  27.90 & 0.827\\
3 & Proposed w/ Source $\ell_1$ loss & 27.14 & 0.807\\
% 5 & Color channel swapping & 28.36  \\
4 & Proposed w/o Source SSIM loss &  29.38 & 0.902\\
5 & Proposed & \textbf{30.30} & \textbf{0.913} \\ \hline
\end{tabular}
}
\end{table}

\textbf{Ablation Study:} \label{sec:ablation}
We discuss relevant ablations for our proposed method trained with Sony as source and 4-shot Nikon as target (Refer Table \ref{tab:ablation}). The details are as follows:
%\begin{itemize}[wide,itemindent=2.5em,noitemsep,topsep=0pt,itemsep=0pt]
\begin{itemize}[wide,itemindent=0.5em,noitemsep,topsep=0pt,itemsep=0pt]
% \item \textit{CIE-XYZ Input}: In the processing of a RAW image to sRGB, there is an intermediate conversion to the CIE-XYZ common color space before applying non-linear post-processing steps. We formulate this experiment as a CIE-XYZ to sRGB conversion to validate the requirement for domain adaptation for the extreme low-light image enhancement task. An LSID model trained with Sony data as source and fine-tuned on 4 Nikon camera images attains a PSNR of 27.16 and SSIM of 0.894 which is lesser than the proposed approach.
\item \textit{Separate Encoder and Decoder}: Transfer learning methods performed sub-optimally primarily due to the color bias. Hence, we trained separate encoders, a shared U-net, and separate camera-specific decoders to learn different camera color spaces individually. This approach obtained a PSNR of 28.62dB and SSIM of 0.867, and the results had visible color gaps between the model’s output and ground truth.
\item \textit{Combined Encoders}: To verify camera-specific denoising by the encoders, we used a shared encoder followed by a shared $\mathbb{N}$ network. But, we noticed the colors of the target domain's output to be dominated by the large source domain data. We obtain a PSNR of 29.20dB and SSIM of 0.890, which is better than fine-tuning LSID, but the colors of the target domain’s output validation images were dominated by the source training images.
% \item \textit{LAB input}: We trained the proposed model using input images in the LAB color space to investigate color quality performance. We used separate models to train for the L-channel and the a,b-channels. The $L_1$ and $\mathcal{L}_{SSIM}$ loss are used for the L-channel; only $L_{1}$ loss was used for the a,b-channels. We obtain L and ab outputs separately from the model during inference time and convert them to the corresponding RGB image. The outputs showed higher color loss and an increase in noise obtaining a PSNR of 27.90dB and SSIM of 0.827.
\item \textit{Proposed w/ Source $\mathcal{\ell}_{1}$}: Training with a strong supervision loss for the source domain strongly influenced the colors of the target domain's output images.
\item \textit{Proposed w/o $\mathcal{L}_{SSIM}$}: From previous ablations, we observe the source domain influencing the target output image color. Hence, we separate only the encoder while merging the U-net and the decoder to get ($\mathbb{N}$). We experimented with several loss functions, including combinations of $\ell_{2}$ loss, grayscale SSIM, gradient loss, $\ell_1$ loss, and cosine similarity loss. We found that using cosine similarity loss and SSIM loss for the source domain, and $\ell_{1}$ loss \cite{zhao2017loss} for the target led to better preservation of color and structure information. 
\item \textit{The $\mathcal{D}$ network for SSIM loss}: As discussed in section \ref{sec:method}, type-casting the source output from 16-bit to 8-bit space will still possess domain specific details. The SSIM loss is used to compare the brightness and structural details but not the color quality (cosine similarity is for color). Thus, following the camera ISP processing steps, we train a 16-to-8-bit conversion U-net model ($\mathcal{D}$ in Fig. 2(b)) to convert the 16-bit data to 8-bit data. The $\mathcal{D}$ network is trained to perform the following non-linear operations: White balancing, Gamma correction, Quantization and JPEG compression. From experiments, we observe that a U-net is necessary to learn all the above mentioned non-linear operations. 
% \item \textit{The $\mathcal{D}$ network and SSIM loss.} As discussed in section \ref{sec:pipeline}, type-casting the source output from 16-bit to 8-bit space will still possess domain specific details. The SSIM loss is used to compare the brightness and structural details but not the color quality (cosine similarity is for color). Thus, following the camera ISP processing steps, we train a 16-to-8-bit conversion U-net model ($\mathcal{D}$ in Fig. \ref{fig:prop_overview}) to convert the 16-bit data to 8-bit data. The $\mathcal{D}$ network is trained to perform the following non-linear operations: White balancing, Gamma correction, Quantization and JPEG compression. 
\end{itemize}

\vspace{-0.5cm}\section{Conclusion}
\label{sec:conclusion}
    In this work, we propose a novel domain adaptation method for low-light raw image enhancement using only a few labeled samples from the target domain and many source domain samples. We first extract camera-specific information through separate encoders, and then use a shared enhancement network ($\mathbb{N}$) to extract domain invariant features and transfer the task successfully from the source domain to the target domain. We propose to compute post-processed image loss for the source domain to have less influence on predictions in the target domain. We also present a new labeled raw image dataset captured with a Nikon camera. Our results show that using only a few labeled samples from the target domain is sufficient to obtain similar or better results than training with large target domain data. We hope that our method and dataset inspire new investigations along these research directions. 

% Then, we use a shared enhancement network to perform denoising, color enhancement, and demosaicing together for both the source and target domains. % the research community to further
    % % use section* for acknowledgment
% \vspace{-0.5cm}\subsubsection*{Acknowledgment}\vspace{-0.3cm}
\noindent\textbf{Acknowledgment}: This work was supported by a project grant from MeitY (No.4(16)/2019-ITEA), Govt. of India.
\bibliography{main}
\end{document}